\documentclass[letterpaper, 10 pt, conference]{ieeeconf}

\IEEEoverridecommandlockouts     

\usepackage{graphicx}
\usepackage{subcaption} 
\usepackage{float}

\usepackage{stfloats}
\usepackage{amsmath} 
\usepackage{url}
\usepackage[bottom,symbol]{footmisc}
\usepackage{balance}

\title{\LARGE \bf
Enabling a Pepper Robot to provide Automated and Interactive Tours of a Robotics Laboratory
}

\author{Gavin Suddrey$^{1}$, Adam Jacobson$^{1}$ and Belinda Ward$^{1}$
\thanks{$^{1}$Gavin Suddrey, Adam Jacobson and Belinda Ward are from the Robotics and Autonomous Systems Disipline, within the Science and Engineering Faculty, Queensland University of Australia, Queensland, Australia.}}

\begin{document}

\onecolumn
\noindent\textcopyright 2018 IEEE. This work has been submitted to the IEEE International Conference on Intelligenct Robots for possible publication. Copyright may be transferred without notice, after which this version may no longer be accessible.

\twocolumn

\newpage
\maketitle
\thispagestyle{empty}
\pagestyle{empty}

\begin{abstract}
The Pepper robot has become a widely recognised face for the perceived potential of social robots to enter our homes and businesses. However, to date, commercial and research applications of the Pepper have been largely restricted to roles in which the robot is able to remain stationary. This restriction is the result of a number of technical limitations, including limited sensing capabilities, and have as a result, reduced the number of roles in which use of the robot can be explored. In this paper, we present our approach to solving these problems, with the intention of opening up new research applications for the robot. To demonstrate the applicability of our approach, we have framed this work within the context of providing interactive tours of an open-plan robotics laboratory. 
\end{abstract}


\renewcommand{\thefootnote}{\fnsymbol{footnote}}

\section{Introduction}
Social robots have recently gained a large degree of attention for their perceived ability to solve many challenges faced by modern society. This interest, which flows from both the research community, and the media, has largely focused on how such robots could be appropriately applied in areas such as healthcare \cite{RABBITT201535}, and education \cite{Tanaka2015}. However, before these robots can be deployed with any degree of success, a number of significant challenges must first be overcome. Amongst these challenges is the ability to accurately navigate within human-centric environments, as well as the need to interact effectively with the people within these environments \cite{10.1007/978-3-319-47437-3_74}. Exploring methods to overcome the challenge of navigating and interacting with people in complex environments, robots such as Minerva \cite{Minerva} and FROG \cite{10.1007/978-3-319-25554-5_32}, have previously been employed within the context of providing autonomous tours of large-scale museums. 

The Pepper robot, a social robot released by Softbank Robotics, has become a well recognised example of social robotics moving into commercial spaces. However, despite Pepper's engaging outward appearance, it has so far been mostly relegated to stationary tasks, such as working as a hotel concierge, or promoting commercial products using its uniqueness to draw in potential customers \cite{lier2018}. This is largely due to limitations in the hardware and software of the robot that make navigating through complex environments a challenging task. These limitations include unpredictable motion of the robot when turning at speeds even below walking speed; a limited sensor suite that make both mapping and localising within its environment a challenge; and a restrictive software development stack that is difficult to utilise with larger off-the-shelf software packages, and in many aspects, largely out-of-date. In order to conduct social robotics experiments with Pepper robots in contexts such as guides, these limitations must be overcome.

In this paper, we describe our approach to solving the problems which we have previously outlined, and demonstrate how the application of our approach to a Pepper robot enables it to provide autonomous and interactive tours of an open-plan robotics laboratory. To summarise, the contributions\footnote{All software resources developed as a result of this work have been open sourced and can be downloaded from from https://bitbucket.org/pepper\_qut} which we present in this paper include:
\begin{enumerate}
\item An up-to-date virtual machine image environment in which complex software can be compiled and deployed directly to the robot.
\item A novel motion controller that resolves issues with the unpredictable motion of the robot when executing turns.
\item A method for localising and navigating through complex open-plan environments using Pepper's limited sensor suite.
\item ROS/NAOQi bindings that enable the NAOQi software, the underlying software that controls the Pepper robot, to take advantage of the ROS navigation stack.
\item A method for generating usable 2D metric maps directly from 3D point clouds that enables the Pepper robot to accurately localise within the environment.
\end{enumerate}

\begin{figure}
\centering
  \includegraphics[width=0.48\textwidth]{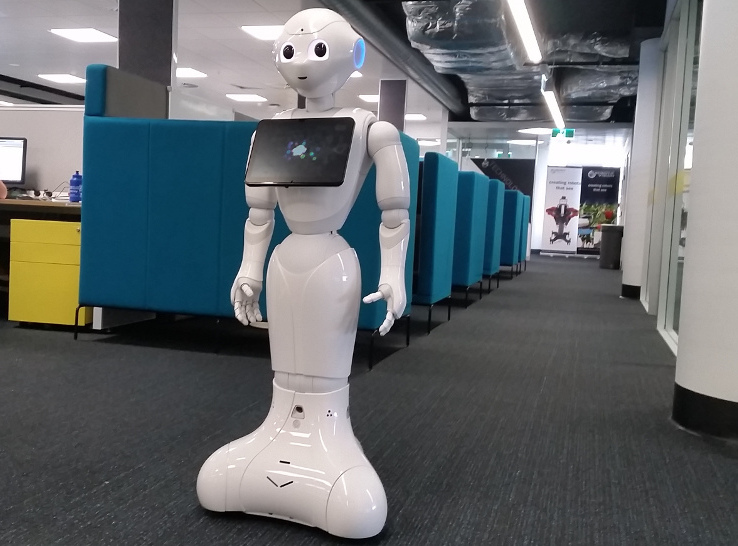}
  \caption{The Pepper robot was able to provide interactive tours by moving through the laboratory and office space. A map of this space can be seen in Figure \ref{fig:map}.}
  
\end{figure}
\section{Related Work}
While the Pepper robot has been popular in media as a potentially game-changing social robot, most prior research involving the robot has concentrated solely on its use as a conversational agent, such as the work described in \cite{10.1007/978-3-319-47437-3_74}. Prior work that has attempted to utilise Pepper as a mobile agent has relied on manual methods of controlling the robot's motion, without having to rely on autonomous navigation techniques. The work presented in \cite{7405071} makes use of an approach by which the robot can be maneurvered by manipulating the position of the robot's end effector with respect to a default pose. Another more traditional approach that is often used in human-robot interaction research involving Pepper has been the use of teleoperation \cite{10.1007/978-3-319-47437-3_74}. 
 
To date, there has been very little work exploring Pepper's ability to autonomously navigate around its environment. In fact, while there is an autonomous navigation software suite available on the robot, it makes use of highly sparse laser scanners located at the base of the robot, and is therefore unable to localise beyond toy examples. To resolve this, the work presented in \cite{Perera17} investigated using laser scans derived from the depth sensor, rather than the laser scanners, to enable the robot to build a map of its environment and localise itself. However, the environment that was mapped in this work was a single small room. For larger open-plan environments, using standard map-building techniques with the depth sensor was found to be extremely error prone.

Other work has looked at providing a novel motion controller for the Pepper robot, investigating how the upper-body of the robot could be used to provide increased stability to the robot when moving at speed around corners \cite{Lafaye2014}. However, we note with interest that this work makes no specific mention of the unpredictable behaviour of the robot when turning at speeds below walking speed. 

\section{Approach}
The following section outlines and describes the approach taken in order to enable a Pepper robot to provide an autonomous and interactive tour of an open-plan robotics laboratory and office area. This approach involved a number of solutions to problems identified with the platform, including a difficult to use and largely out-dated software development stack; a limited sensor suite; and unpredictable motion of the robot when executing turns.

\subsection{Development Environment}
The currently perscribed approach to using ROS within the ROS documentation for the Pepper robot is to compile and run any ROS based packages from a remote machine, such as a desktop, which then communicates with the Pepper robot over a wireless network. However, for computationally intensive tasks such as localisation and navigation, the latency introduced by offboard processing introduces numerous issues, and wireless network dropouts while the robot is moving can be potentially disasterous. To avoid this issue, previous works have described steps for installing ROS on a Pepper robot, that have utilised the official virtual machine image compiled by Softbank Robotics to build ROS \cite{Perera17}. However, this package is no longer actively developed by Softbank Robotics, with the most recent version dating back to 2014. Due to version differences between the virtual machine available from Sotbank Robotics, and the operating system of the robot, these approaches are no longer viable. 

To address this issue, we have developed a custom linux virtual machine image that has been designed to mirror the most recent version to date of the Pepper robot's operating system - NAOQi OS 2.5.5. With this virtual machine, we have been able to build the necessary dependencies and ROS packages to provide the Pepper robot with complete access to the ROS Navigation Stack. The current ROS distribution in use on the robot is ROS Indigo, but we are exploring the creation of a ROS Kinetic build with the use of the virtual machine.

\subsection{Improved Motion Controller}

\begin{figure}[t]
    \centering
    \begin{subfigure}[b]{0.48\linewidth}
        \includegraphics[width=\textwidth]{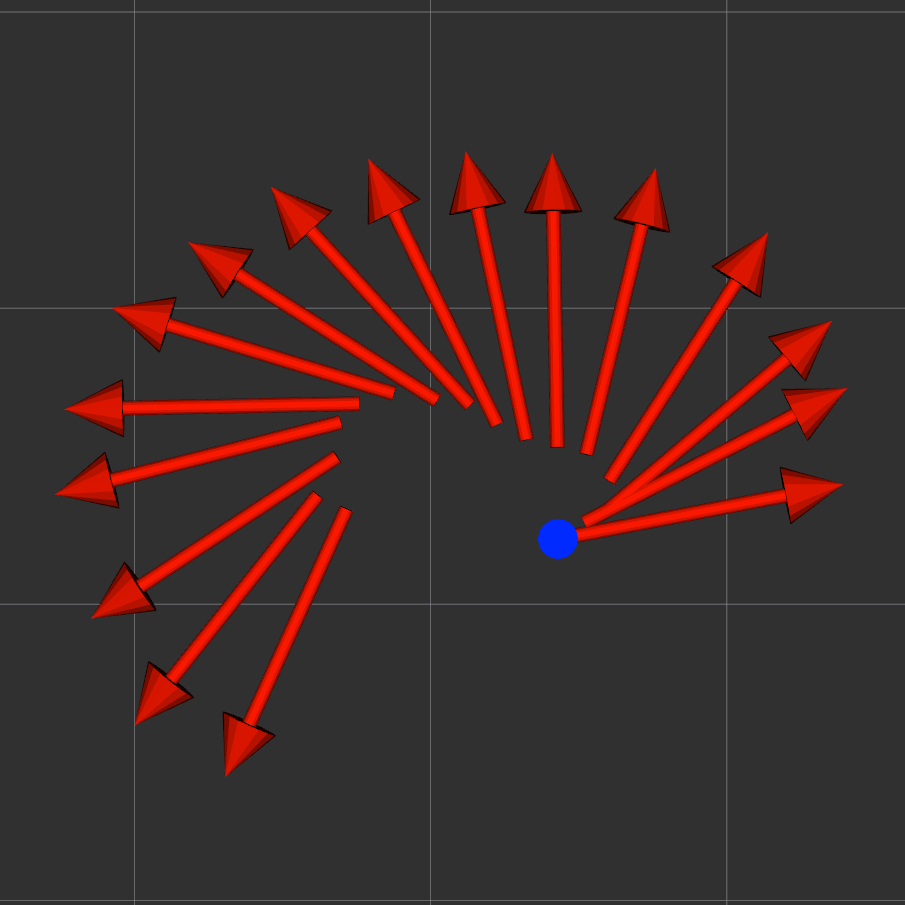}
        \caption{Vanilla Controller}
        \label{fig:controller_vanilla}
    \end{subfigure}
        \hfill 
     \begin{subfigure}[b]{0.48\linewidth}
        \includegraphics[width=\textwidth]{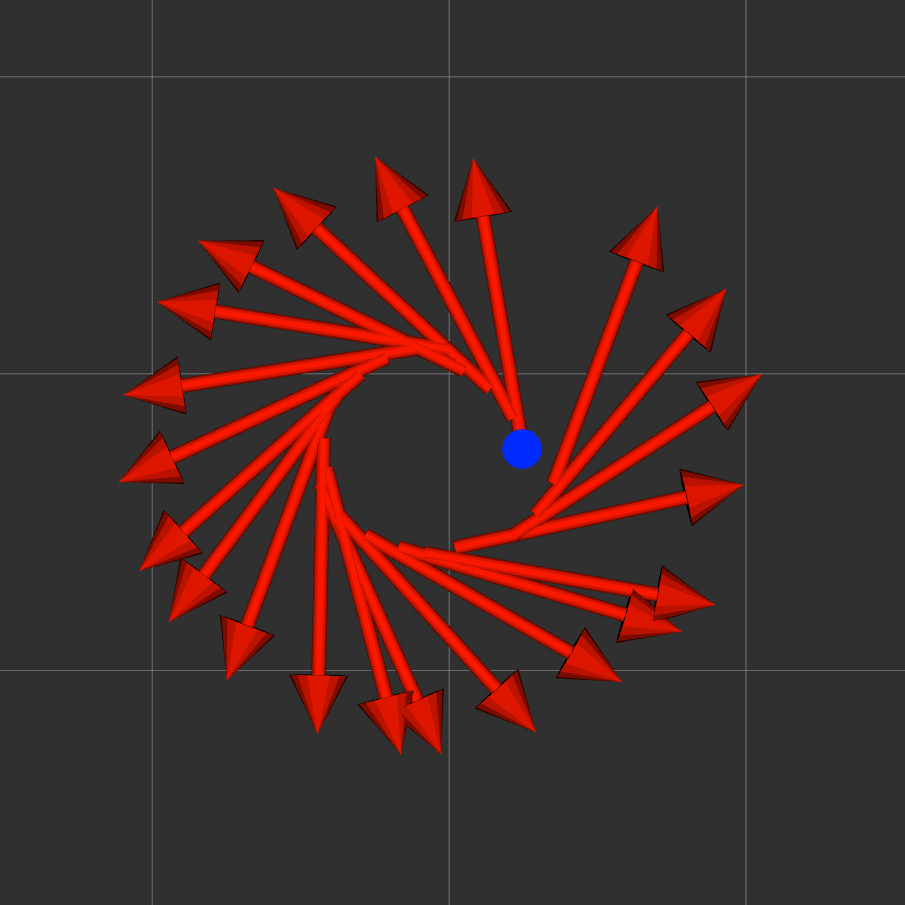}
        \caption{Corrected Controller}
        \label{fig:controller_corrected}
    \end{subfigure}
    \caption{The vanilla motion controller packaged with the robot generates unexpected motion when trying to drive in a circle as seen in Figure \ref{fig:controller_vanilla}. To correct this, a custom controller was implemented, which resulted in more predictable motion from the robot, as seen in Figure \ref{fig:controller_corrected}. Note that the blue dot indicates the starting point of the motion.}
\end{figure}

\begin{figure}[H]
\centering
  \includegraphics[width=0.48\textwidth]{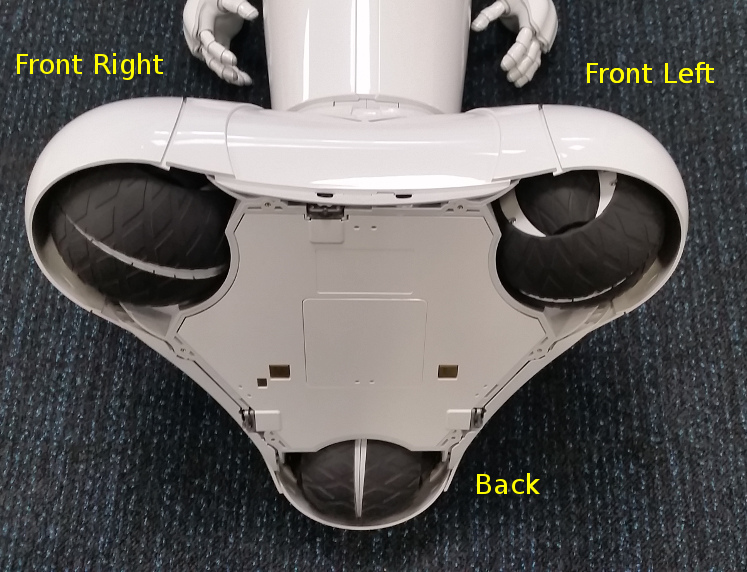}
\caption{Pepper is a holonomic robot with three omni-directional wheels that function together to generate motion.}
  \label{fig:wheels}
\end{figure}

Interestingly, we found that the motion of the robot when turning along the circumference of a circle at speeds even below walking speed were unpredictable. As can be seen in Figure \ref{fig:controller_vanilla}, rather than facing along the direction of travel, the robot would instead face outward from the centre of rotation. In order to correct the motion of the robot, a custom motion controller was implemented that would correctly account for the holonomic nature of the base of the robot. The equations used to create this controller are described in \cite{ribeiro2004three}, where the motion of a holonomic platform can be described using the the following equations:
\begin{equation}
\label{eq:wfl}
W_\text{fl} = \theta + V \cdot \text{cos}(150 - \omega)
\end{equation}
\begin{equation}
\label{eq:wfr}
W_\text{fr} = \theta + V \cdot \text{cos}(30  - \omega)
\end{equation}
\begin{equation}
\label{eq:wb}
W_\text{b} = \theta + V \cdot \text{cos}(270 - \omega)
\end{equation}

\begin{figure*}[t]
    \centering
    \begin{subfigure}[b]{0.32\linewidth}
        \includegraphics[width=\textwidth]{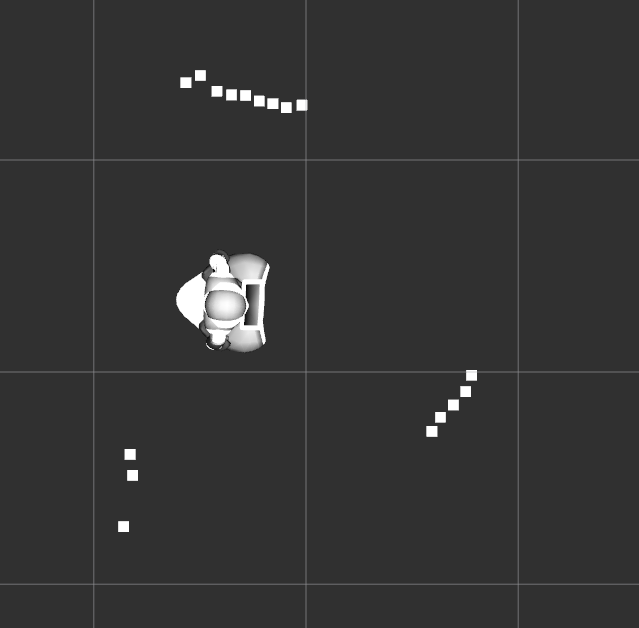}
        \caption{Scan from Base Lasers}
        \label{fig:scan_base}
    \end{subfigure}
        \hfill 
     \begin{subfigure}[b]{0.32\linewidth}
        \includegraphics[width=\textwidth]{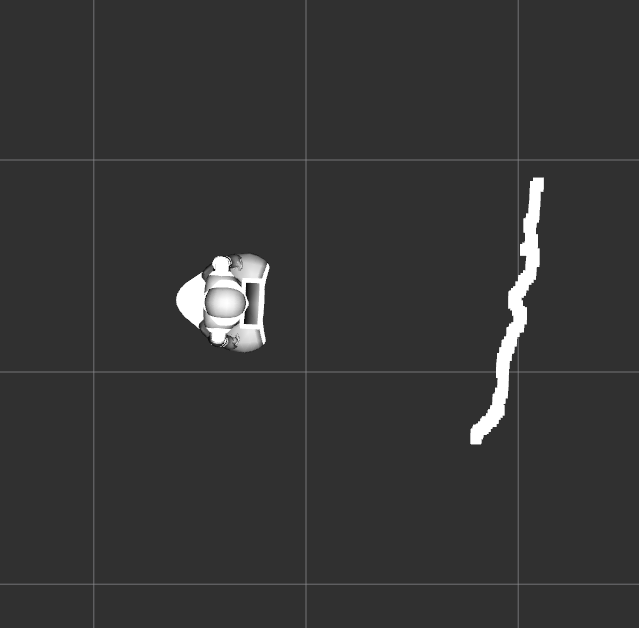}
        \caption{Scan from Depth Sensor}
        \label{fig:scan_depth}
    \end{subfigure}
        \hfill 
     \begin{subfigure}[b]{0.32\linewidth}
        \includegraphics[width=\textwidth]{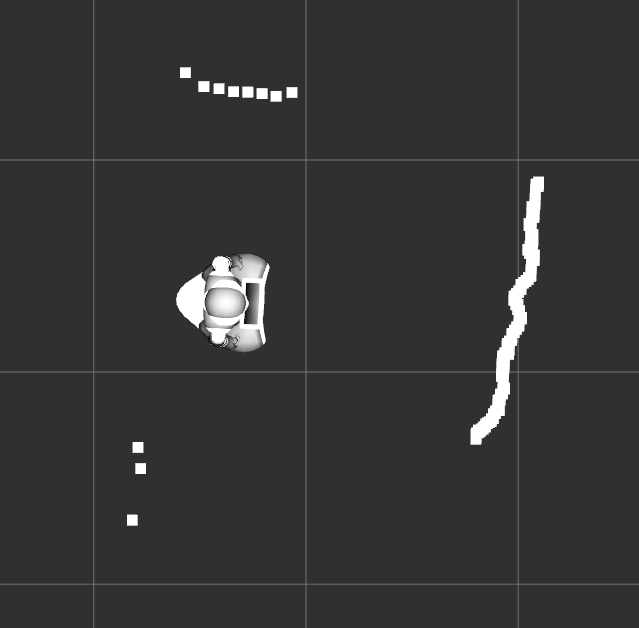}
        \caption{Resulting Merged Scan}
        \label{fig:scan_merged}
    \end{subfigure}
    \caption{The point sparsity of the base lasers can be observed in Figure \ref{fig:scan_base}. The laser scan, extracted from the depth image, is more dense but constrained to a narrow field of view as seen in Figure \ref{fig:scan_depth}. The merged scan can be seen in Figure \ref{fig:scan_merged}. The scanned environment is visible in Figure \ref{fig:obstacles}.}
    \label{fig:scans}
\end{figure*}

\begin{figure}
\centering
  \includegraphics[width=0.5\textwidth]{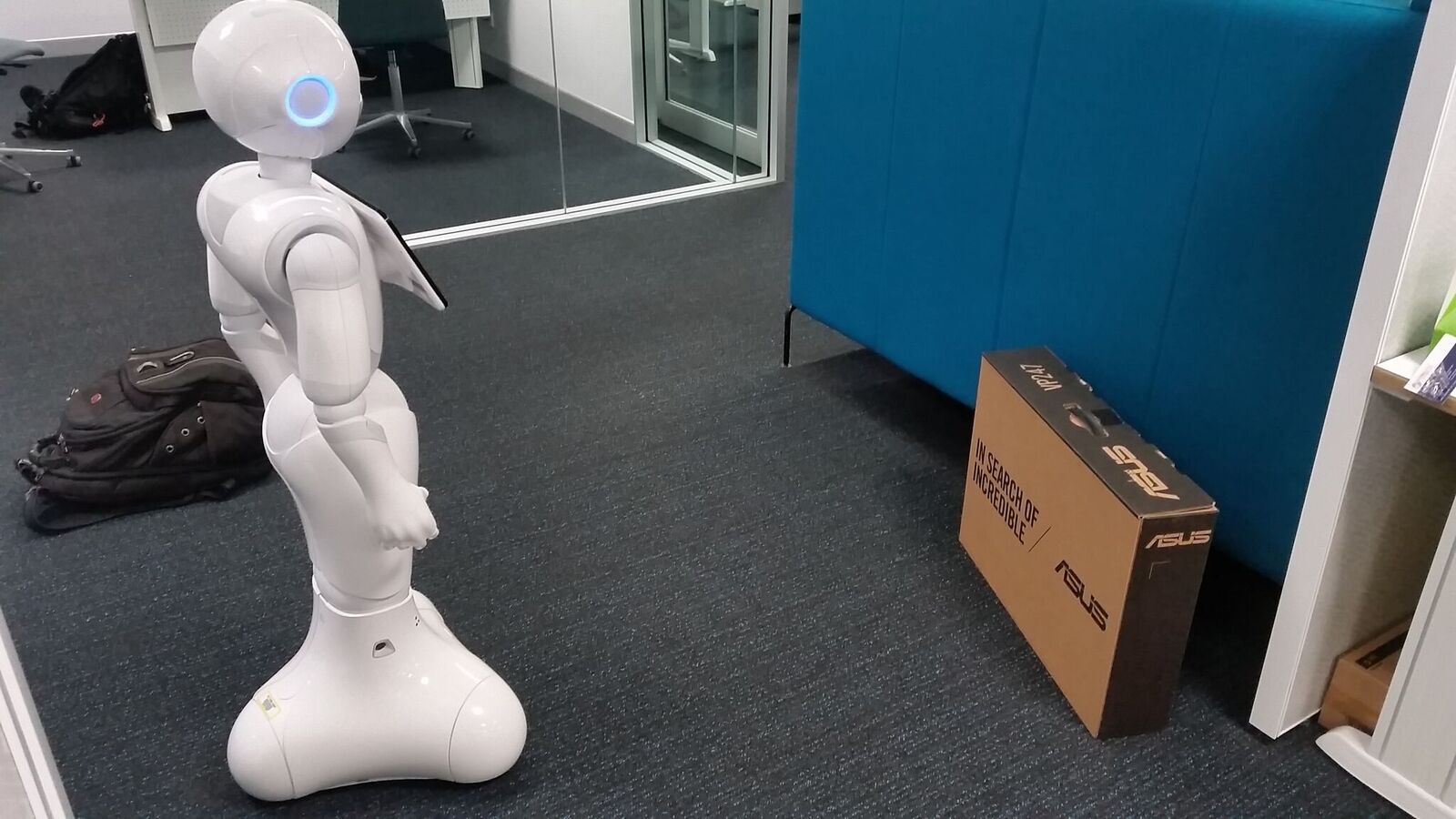}
\caption{The obstacles detected within Figure \ref{fig:scans}. It is worth noting that the base lasers in Figure \ref{fig:scan_base} were not able to pick up the sofa directly ahead, due to its elevation above the ground. Conversely, the scan from the depth sensor in Figure \ref{fig:scan_depth} was unable to pick up the backpack and the box.}
  \label{fig:obstacles}
\end{figure}

where $W_\text{fl}$, $W_\text{fr}$, and $W_\text{b}$ represent the front-left, front-right and back wheels respectively (see Figure \ref{fig:wheels}); $\omega$ describes orientation of the drive vector of the robot; $V$ describes the magnitude of the drive vector; and $\theta$ represents an angular rotational value that is added equally across all three wheels. For positive values of the rotation component, the robot rotates to the left, while negative values rotate the robot to the right. Using this approach, we were able to correct the motion of the robot, as can be seen in Figure \ref{fig:controller_corrected}.

\subsection{Localisation and Navigation}
\subsubsection{Localisation}
\begin{figure}[t]
\centering
\begin{subfigure}[b]{1\linewidth}
\includegraphics[width=1\textwidth]{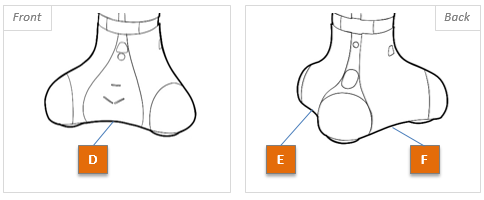}
  \end{subfigure}
  \begin{subfigure}[b]{1\linewidth}
      \includegraphics[width=1\textwidth]{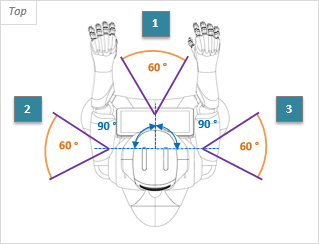}
  \end{subfigure}
  \caption{The Pepper robot has three lasers located at the base of the robot. Each laser has a field-of-view of 60 degrees, and are non-overlapping, creating blind spots during obstacle avoidance. Additionally, each laser provides only 15 points, making them insufficient for localisation. Source \protect\url{http://doc.aldebaran.com/2-5/family/pepper_technical/laser_pep.html}}
  \label{fig:lasers}
\end{figure}

For localisation, we made use of the ROS implementation of the Adaptive Monte Carlo Localisation algorithm \cite{DFox}. This approach utilises the odometry of the robot as well as laser scans of the surrounding environment to generate a particle filter describing the probable poses of the robot. However, while the odometry was found to be accurate over short distances, the sparsity of the laser scans generated by the three laser scanners located at the base of the robot (see Figure \ref{fig:lasers}) provided insufficient resolution (15 points per scanner) to perform localisation (see Figure \ref{fig:scan_base}). 

To obtain denser laser scans, we followed the approach described in \cite{Perera17}, where a simulated laser scan is extracted from a horizontal slice of the depth image (see Figure \ref{fig:scan_depth}). Expanding on this approach we also then merge the laser scans obtained from the depth sensor, which is located in the head of the Pepper robot, and the base laser to generate a more complete scan of the surrounding area (see Figure \ref{fig:scan_merged}). This merged scan was created by generating a new scan area that encompassed both the base and depth scans, with the resolution of the new scan area being equivalent to that of the depth scan. Both scans are then overlayed over the new scan area, with the closest point being selected in areas of overlap. To account for the sparser resolution of the base scan, any regions of the new scan area that fall between the points of the base scan are then set to negative values to indicate invalid results.

\subsubsection{Navigation}
Navigation was achieved using the ROS Move Base package, which provides access to a number of tools for performing real-time navigation in dynamic environments. 

\begin{figure}
\centering
  \includegraphics[width=0.5\textwidth]{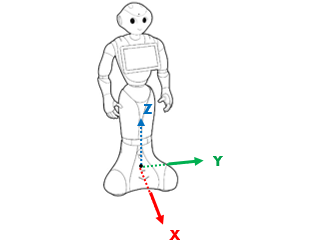}
  \caption{The coordinate frame used during motion of the robot. Adapted from \protect\url{https://android.aldebaran.com/sdk/doc/pepper-sdk/objects/frame.html}}
  \label{fig:frame}
\end{figure}

While we have access to a merged scan of the surrounding region, the sparsity of the base lasers, as well as their extremely low profile, meant that we could not guarantee that the obstacle was no longer present, or if it was simply being missed. To account for this issue, the local costmap used in local path-planning and obstacle avoidance could be populated by points detected in the merged laser scan; obstacles could only be cleared from the costmap however, when the robot was facing the direction of the perceived obstacles and they were no longer present, meaning they had disappeared from both sensors. Additionally, the robot was forbidden from moving along the Y-axis, (see Figure \ref{fig:frame}), due to the chance that obstacles might be missed by the base lasers. Based on this approach, the Pepper robot was then able to successfully navigate to any accessible point within the laboratory.

To facilitate ease-of-use of the navigation system by developers more familiar with the NAOQi development framework, a set of bindings have also been created that enable NAOQi to make use of a set of ROS services that enable the developer to direct the robot to pre-defined locations.

\subsection{Generating Useful Maps}
In order to generate the maps needed to localise and navigate the robot, we initially attempted to use the ROS gmapping software suite. However, this turned out to be extremely challenging with the Pepper, with the maps quickly becoming corrupted even over short distances. To combat this, we explored using a map generated by a Pioneer robot equipped with a SICK LMS-200 laser range-finder. While this worked for our initial experiments, the map was already over a year old at this stage, and contained numerous errors with respect to the actual environment. Further, given the height discrepancy between the Pepper robot and the Pioneer, the map had to be manually annotated to add in features that were not visible to the Pioneer.

\begin{figure}[t]
\centering
  \includegraphics[width=0.48\textwidth]{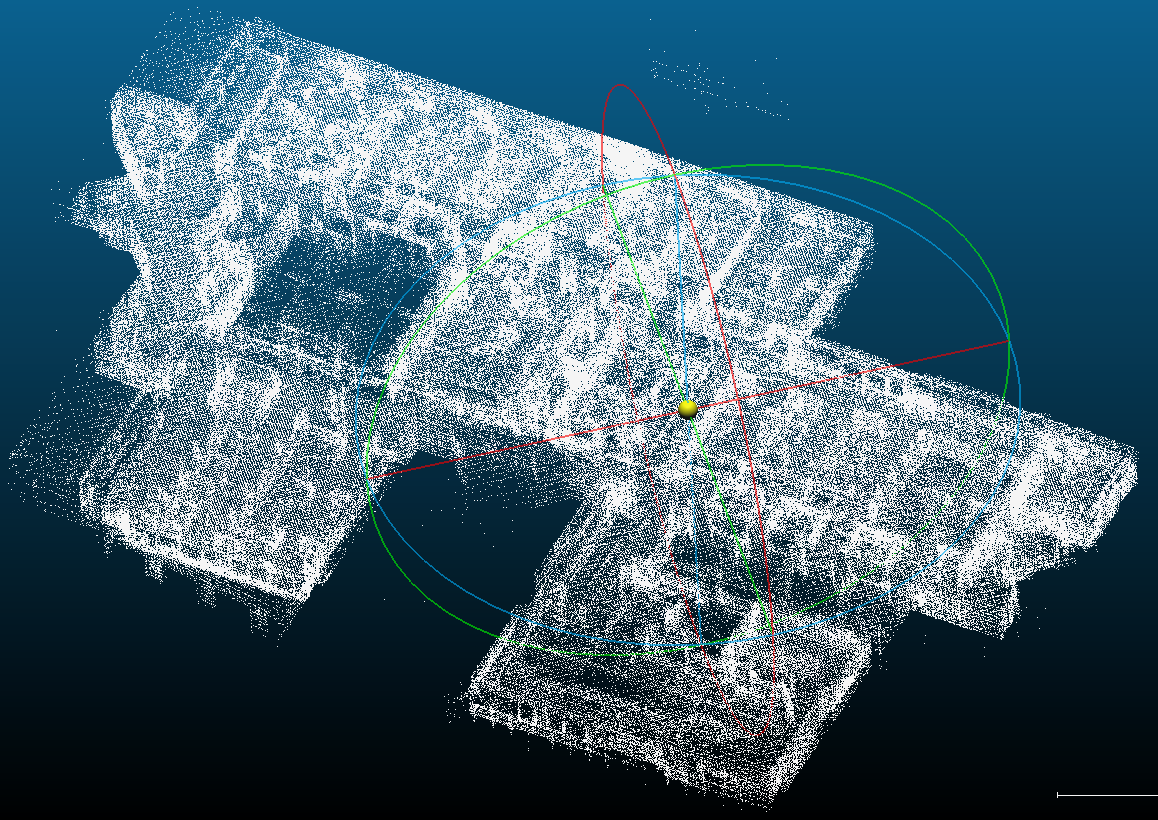}
  \caption{A 3D point cloud of the laboratory captured using CSIRO's handhendl mapping SLAM technology. It should be noted that the pictured point cloud has been reduced in density for illustrative purposes.}
  \label{fig:point_cloud}
\end{figure}

\begin{figure}[t]
\centering
  \includegraphics[width=0.48\textwidth]{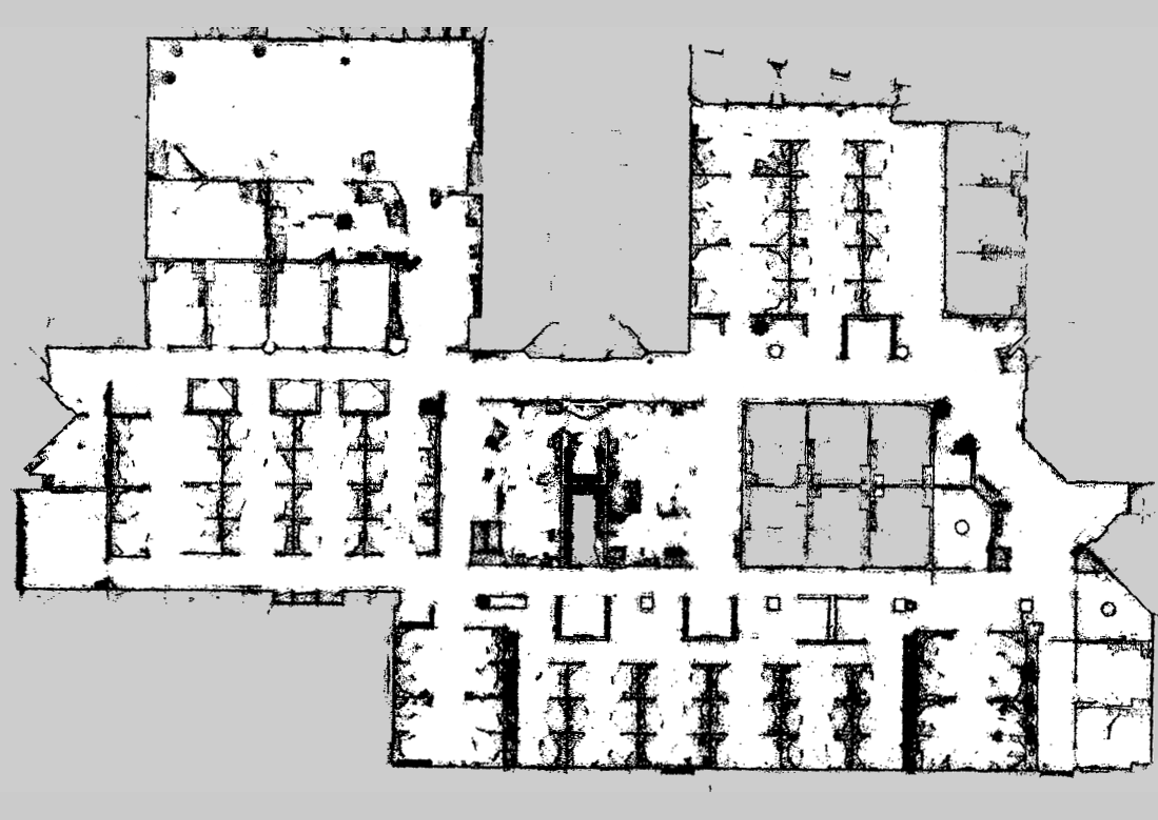}
  \caption{The 2D metric map used by the Pepper robot to localise and path-plan within the labratory. This map was generated from the a denser version of the point cloud seen in Figure \ref{fig:point_cloud}.}
  \label{fig:map}
\end{figure}

In order to obtain more up-to-date maps that could be used by both the Pepper robot, as well as the Pioneer, we then explored the possibility of extracting valid 2D metric maps from a 3D point cloud map of the environment, seen in Figure \ref{fig:point_cloud}. The point cloud maps used in this paper to generate our 2D maps were collected and processed using CSIRO's handheld mobile mapping SLAM technology \cite{revo}, which is commercially availabe through GeoSLAM Ltd.\footnote{GeoSLAM website: \url{https://geoslam.com/}} - a CSIRO joint-venture company. To extract the 2D metric map, the 3D reconstruction was first filtered to only include points located within a certain height range. An output image, representing the 2D map, was then generated, where the width and height of the ouput image corresponded the measured distance in meters between the outer-most points on the both the x and y axis, divided by a fixed ratio describing the correspondance between pixel size and meters. Each pixel in the output image was then set to black if a point existed within the real-world coordinates to which the pixel corresponded, or white if no such point existed. The final output image, which can be seen in Figure \ref{fig:map}, was then post-processed to remove noise, and the outer-region of the map filled to indicate unknown space, which represents unseen space into which the robot should never attempt to drive.

\subsection{Remote Systems}
\begin{figure}
\centering
  \includegraphics[width=0.4\textwidth]{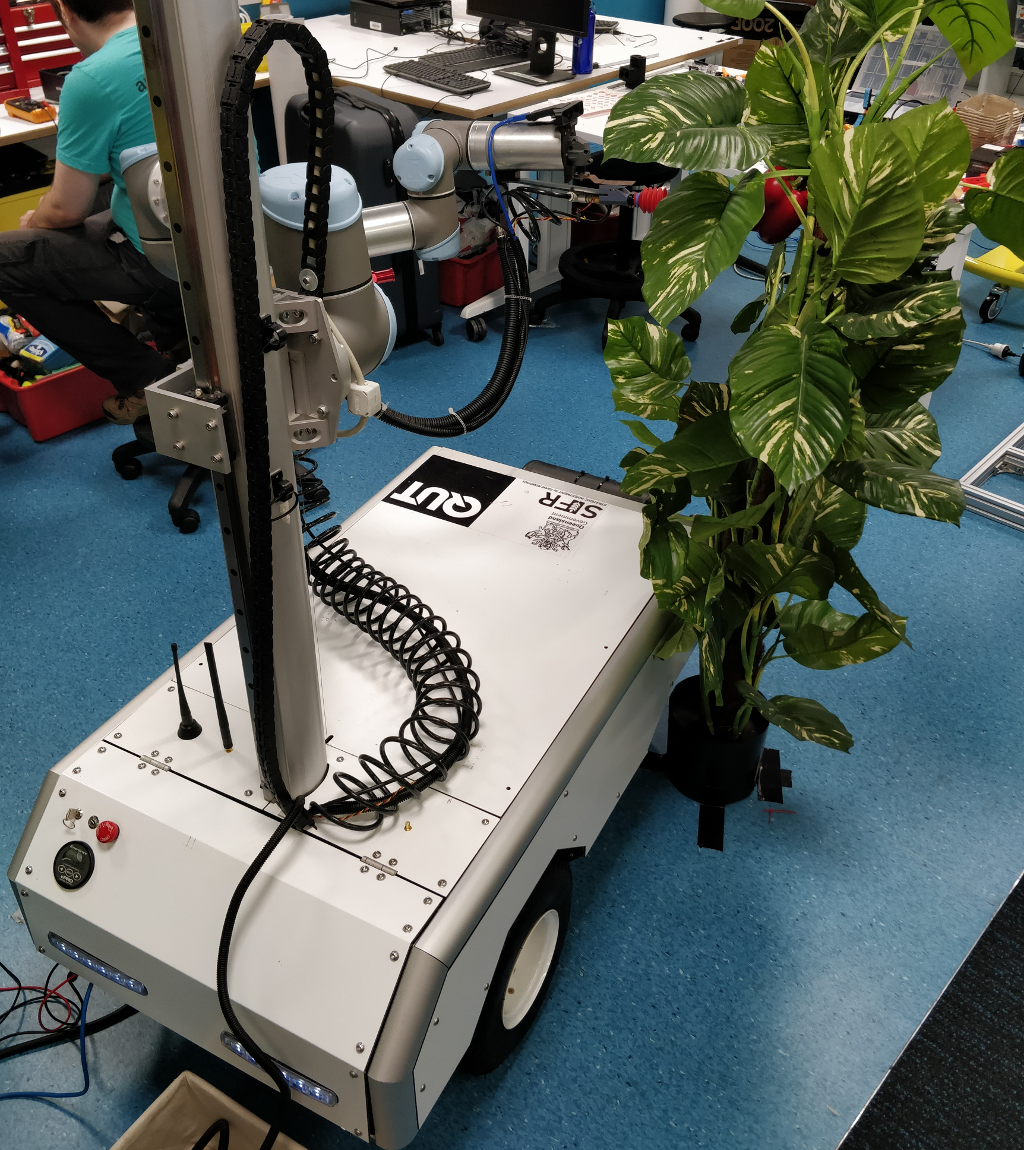}
  \caption{Harvey, resting beside a fake plant with an attached plastic sweet pepper, is a horticultural robot that is designed to use computer vision and its cutting tool to autonomously pick sweet peppers.}
  \label{fig:harvey}
\end{figure}

\begin{figure}
\centering
  \includegraphics[width=0.4\textwidth]{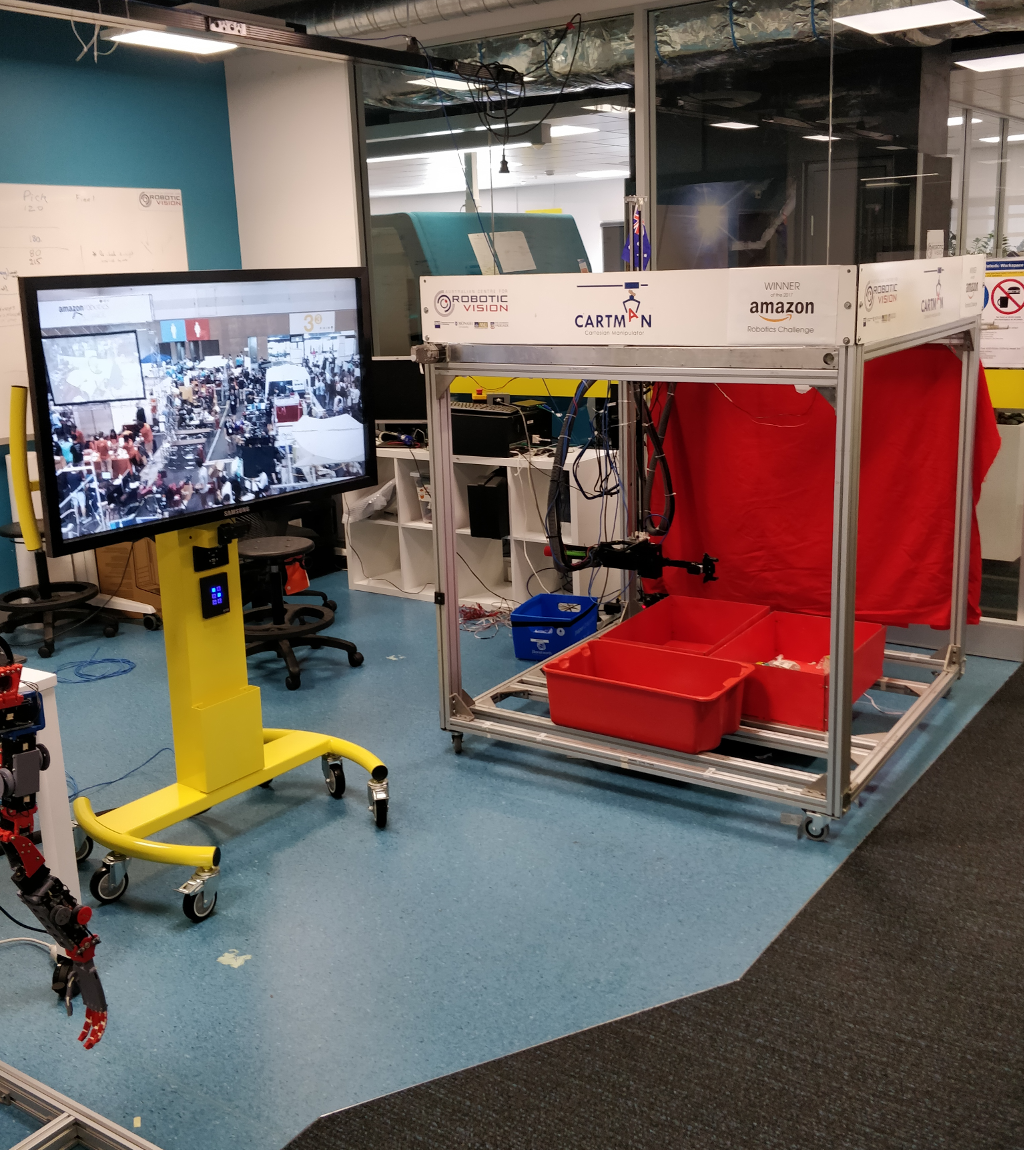}
  \caption{Cartman, the grand final winner of the 2018 Amazon Robotics Challenge, positioned beside the television screen used during the demonstration. Cartman is designed to pick up selected items and place them into shipping boxes.}
  \label{fig:cartman}
\end{figure}

As part of the tour, the Pepper robot was required to interact with a number of other devices within the environment, including two separate robot platforms, Harvey and Cartman, shown in Figures \ref{fig:harvey} and \ref{fig:cartman} respectively. Harvey is a mobile robot that is designed to use robotic vision to identify sweet peppers within dense foliage, and extract an optimal gripping point on the sweet pepper which will allow its cutting tool to remove the fruit from the bush \cite{harvey}. The cartesian manipulator, Cartman, was the winning entrant in the 2018 Amazon Robotics Challenge, and is designed to pick and stow complex, randomly distributed items. In addition to the two robots, the Pepper robot also needed to be capable of interfacing with two large television screens (see Figure \ref{fig:cartman}) equipped with WIFI-enabled computers, positioned next to each robot, in order to play videos relating to the specific robots.

To communicate with each robot, a web server was installed locally on each robot that integrated with the ROS stack of the particular robot. These web servers were capable of starting and stopping various demonstrations, described in Section \ref{sec:demonstrations} on each robot. The Pepper robot could then send simple HTTP requests encoding instructions for individual demonstrations directly to each robot over a shared WIFI network. Each television was also connected to the shared WIFI network in a similar fashion, using a web server that could start, stop and query for media available on the attached computer via HTTP requests made by the Pepper robot.

\section{Results}

\begin{figure*}[t]
\centering
  \includegraphics[width=1.0\textwidth]{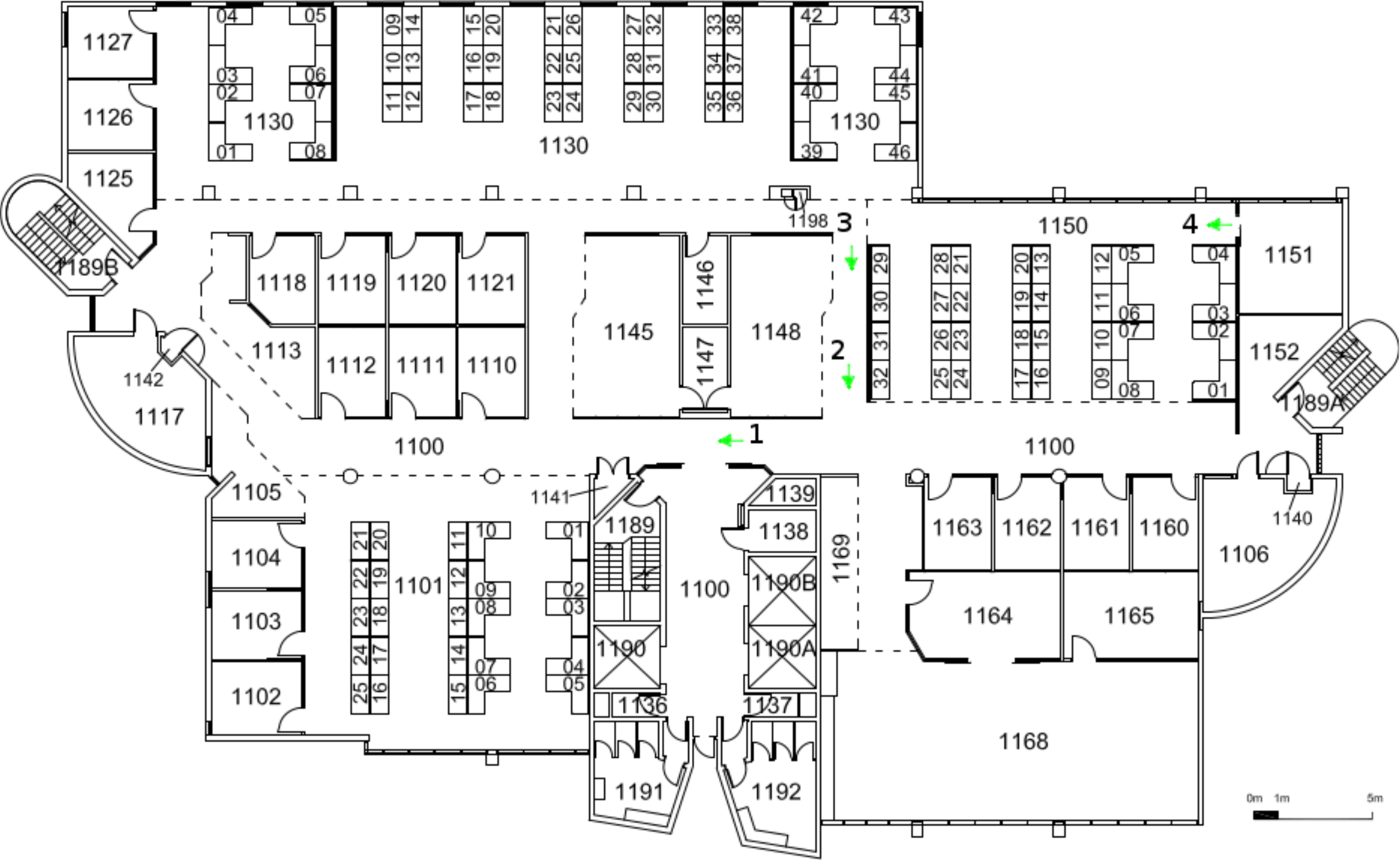}
  \caption{An official floor plan of our laboratory space. Marker 1 indicates the entry to the floor and is where the robot will waits prior to starting to tour; Markers 2 and 3 indicate the Harvey and Cartman stations respectively; and Marker 4 indicates the conference room at which the tour terminates. The vector of each arrow indicates the heading the robot should be in when it has finished navigating to that marker.}
  \label{fig:floorplan}
\end{figure*}

\begin{figure*}[t]
\centering
  \includegraphics[width=1.0\textwidth]{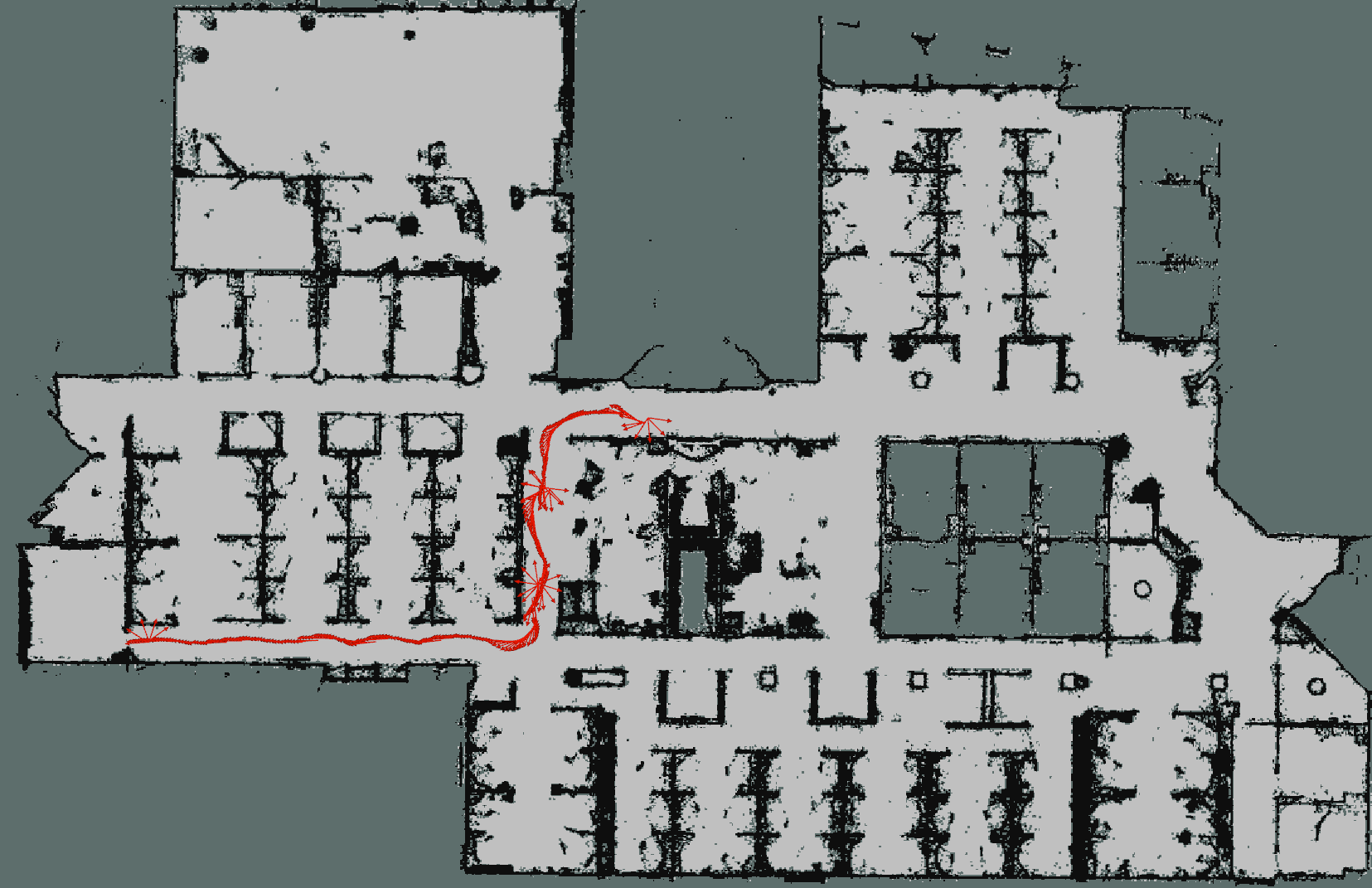}
  \caption{A map highlighting the path taken by Pepper during the course of the tour. The vector of each arrow indicates the heading of the Pepper robot at that point of the tour. The points at which the robot has turned to assume the heading of the makers, as seen in Figure \ref{fig:floorplan}, as well as to resume travelling can be clearly seen.}
  \label{fig:drive_path}
\end{figure*}

To validate our approach, we developed a demonstration in which the Pepper robot would provide autonomous interactive tours of our laboratory. This tour was designed to require the Pepper to move between various points within the environment. An official floor plan of this environment, as well as markers indicating the various points of the tour can be seen in Figure \ref{fig:floorplan}. The odometry shown in Figure \ref{fig:drive_path} shows the path the Pepper has taken to drive between the markers shown in Figure \ref{fig:floorplan}. It should be noted that at each marker the robot was required to turn to face the audience, which can be seen in Figure \ref{fig:drive_path}.  

In addition to driving between the indicated markers, the Pepper robot was also required to deliver interactive demonstrations at Markers 2 and 3. At each of these Markers, Pepper was successful in delivering the relevant demonstration. We describe the demonstrations at these markers, as well as the other interactions at the other two markers below. It should be noted that in addition to this paper, we have also provided a video that demonstrates the robot providing the described tour.

\subsection{Laboratory Tour Description}
\label{sec:demonstrations}

\subsubsection{Marker 1 - Entry}
Marker 1 indicated the beginning of the tour, and was located at the entry way to the laboratory (see Figure \ref{fig:floorplan}).  The robot would remain idle at this marker, displaying both a message on the tablet on its chest welcoming visitors, and a button to begin the tour. When the button was pressed, the robot would check its WIFI connectivity to ensure it could communicate with the demonstrator technologies used in the tour, and then navigate to Marker 2.

\subsubsection{Marker 2 - Harvey Station}
Marker 2 was situated at the Harvey station, and can be seen in Figure \ref{fig:floorplan}. Upon arriving at this marker, the robot would indicate for the tour group to view the television screen directly to its left. The robot would then instruct the television to begin playing a video showing Harvey at working picking capsicums within a greenhouse. Whilst the video was playing, the Pepper would give a detail description of the robot to the tour. Once the video had ended, the Pepper would then instruct the audience to view the Harvey robot, which it would then command to starts it demonstration. This demonstration consisted of the Harvey robot picking a plastic sweet pepper from a fake potted plant located beside it. While the demonstration was in progress, The Pepper would move into an idle state, and display a button on its tablet indicating that it should move on to the next marker of the tour. When this button was pressed, the Pepper would proceed to navigate to Marker 3.

\subsubsection{Marker 3 - Cartman Station}
Marker 3 is located at the Cartman station of the tour. After arriving at this marker, the robot would instruct the audience to view a television screen located next Cartman, as seen in Figure \ref{fig:cartman}, which it would command to play a video of Cartman and the Amazon Robotics Challenge. While this video played, the Pepper would provide a description of Cartman and its participation in the Amazon Robotics Challenge. At the conclusion of the video, the Pepper would then communicate with Cartman to begin its demonstration. This demonstration was interactive, and the Pepper would indicate to the audience to select items from a list on its tablet. When an item was selected, the Pepper would instruct Cartman to pick the selected item, and move it into the receiving bin. This would continue until a finish button was selected, at which stage the robot would proceed to navigate to Marker 4, the end of the tour.

\subsubsection{Marker 4 - Meeting Room}
After arriving at Marker 4, located at a conference room representing the end of the tour, the Pepper robot would indicate for the audience to move into the room.

\section{Conclusion}
In this paper we present a number of contributions aimed at solving the problem of allowing a Pepper robot to navigate autonomously, and provide interactive tours of an open-plan laboratory space. These contributions included a set of tools for facilitating software development on the robot; an updated motion controller; an approach to working with the limited sensor suite available on the robot; and an approach to generating viable maps for navigating within the environment. Additionally, we have outlined how we integrated the Pepper robot with various technologies located within the environment to provide a more interactive experience, including interfacing with and starting two complex robotic platforms.

During the course of this work, we encountered two issues that continue to present problems with rolling the tour out as a regular attraction. The first issue is the dependence on the local WIFI network, which is at times extremely unreliable for the purposes of our demonstration. The second issue is the complexity of the two auxilliary robots used in the tour, Harvey and Cartman, both of which require a certain degree of expert knowledge to simply start up. Additionally, both Harvey and Cartman are active research platforms, and as such are frequently unavailable for tours.

In future work we plan to investigate how robust the current navigation setup is to running the Pepper robot continuously over a long period of time. Additionally, we plan to explore how peoples' attitudes toward the robot are affected by its ability to navigate, when compared to the more traditional setting where the robot is largely stationary. Lastly, we plan to begin using Pepper in a wider variety of human-robot interaction scenarios, such as in health care, that rely more heavily on access to functional navigation capabilities.
\balance

\section*{ACKNOWLEDGMENT}
This work was funded by the Queensland Government under an Advance Queensland Grant.

\bibliographystyle{IEEEtran}  
\bibliography{bibliography}

\end{document}